\documentclass[]{fairmeta}

\usepackage{graphicx} 
\usepackage{amsmath}
\usepackage{xcolor}

\usepackage[utf8]{inputenc} 
\usepackage[T1]{fontenc}    
\usepackage{hyperref}       
\usepackage{url}            
\usepackage{booktabs}       
\usepackage{amsfonts}       

\usepackage{nicefrac}       
\usepackage{microtype}      
\usepackage{multirow}
\usepackage{xcolor}         
\usepackage{cleveref}       
\usepackage{comment}

\usepackage{colortbl}   
\usepackage{array}      

\title{Learning to Reason in 13 Parameters}

\author[1,2]{John X. Morris}
\author[1,3]{Niloofar Mireshghallah}
\author[1]{Mark Ibrahim}
\author[1]{Saeed Mahloujifar}

\affiliation[1]{FAIR at Meta}
\affiliation[2]{Cornell University}
\affiliation[3]{Carnegie Mellon University}

\newtoggle{showcomments}
\toggletrue{showcomments}

\newcommand{\definecommenter}[2]{%
  \newcommand{#1}[1]{%
    \iftoggle{showcomments}{%
      \textcolor{#2}{{{\detokenize{#1}}}: \textit{##1}}%
    }{}%
  }%
}

\definecommenter{\niloofar}{teal}
\definecommenter{\Mark}{purple}

\date{\today}

\abstract{
    Recent research has shown that language models can learn to \textit{reason}, often via reinforcement learning. Some work even trains low-rank parameterizations for reasoning, but conventional LoRA cannot scale below the model dimension. We question whether even rank=1 LoRA is necessary for learning to reason and propose TinyLoRA, a method for scaling low-rank adapters to sizes as small as one parameter. Within our new parameterization, we are able to train the 8B parameter size of Qwen2.5 to 91\% accuracy on GSM8K with only 13 trained parameters in bf16 (26 total bytes). We find this trend holds in general: we are able to recover 90\% of performance improvements while training $1000x$ fewer parameters across a suite of more difficult learning-to-reason benchmarks such as AIME, AMC, and MATH500. Notably, we are only able to achieve such strong performance with RL: models trained using SFT require $100-1000x$ larger updates to reach the same performance.
}
\begin{document}

\maketitle

\begin{figure}
    \centering
    \begin{minipage}{0.48\textwidth}
        \centering
        \includegraphics[width=\linewidth]{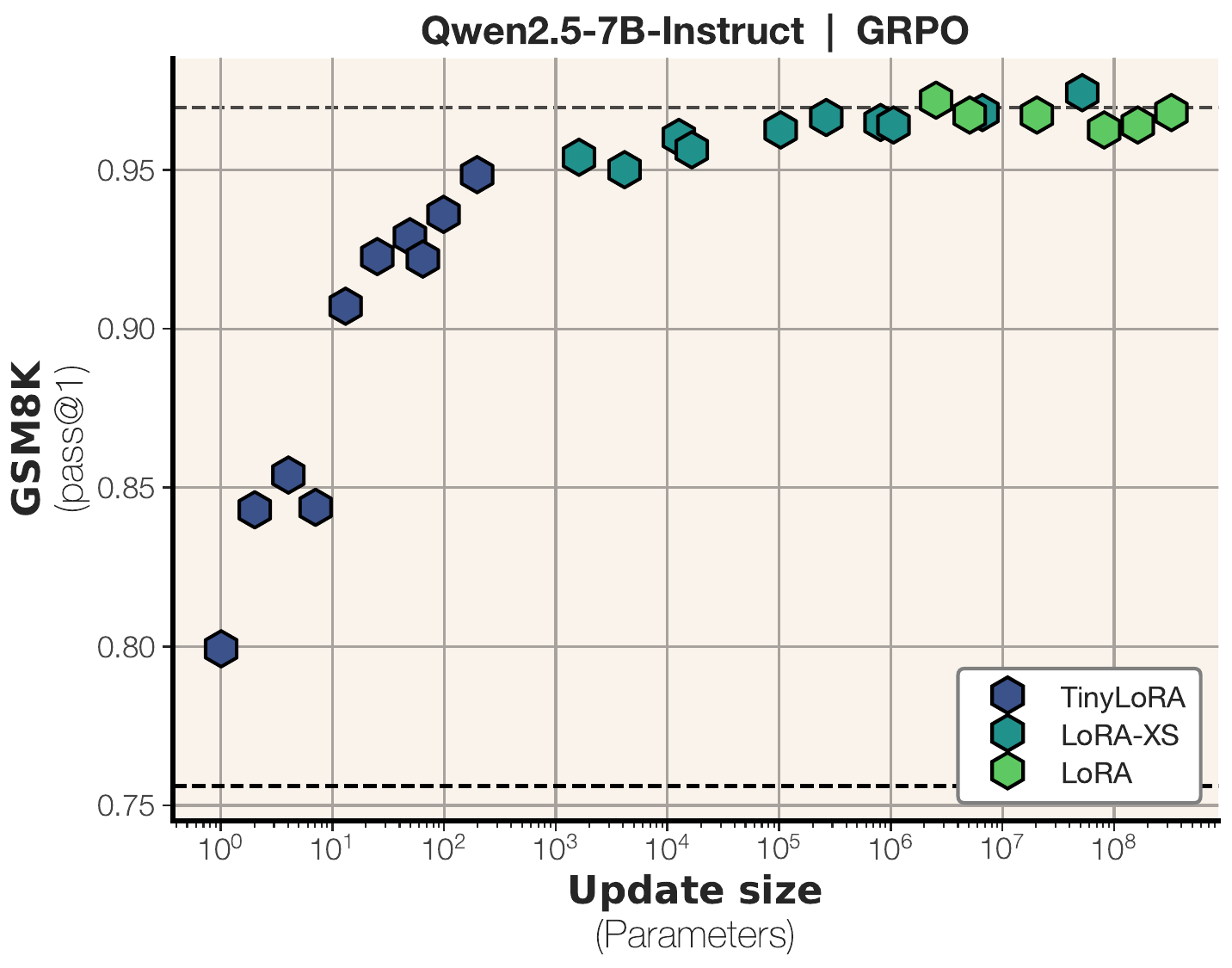}
        \caption{Using Qwen2.5-7B-Instruct as a base model, our TinyLoRA achieves performance within 5\% of full finetuning on GSM8K with only 13 parameters. Dashed lines indicate untrained and full-FT baselines.}
        \label{fig:main-gsm8k-grpo}
    \end{minipage}
    \hfill
    \begin{minipage}{0.48\textwidth}
        \centering
        \includegraphics[width=\linewidth]{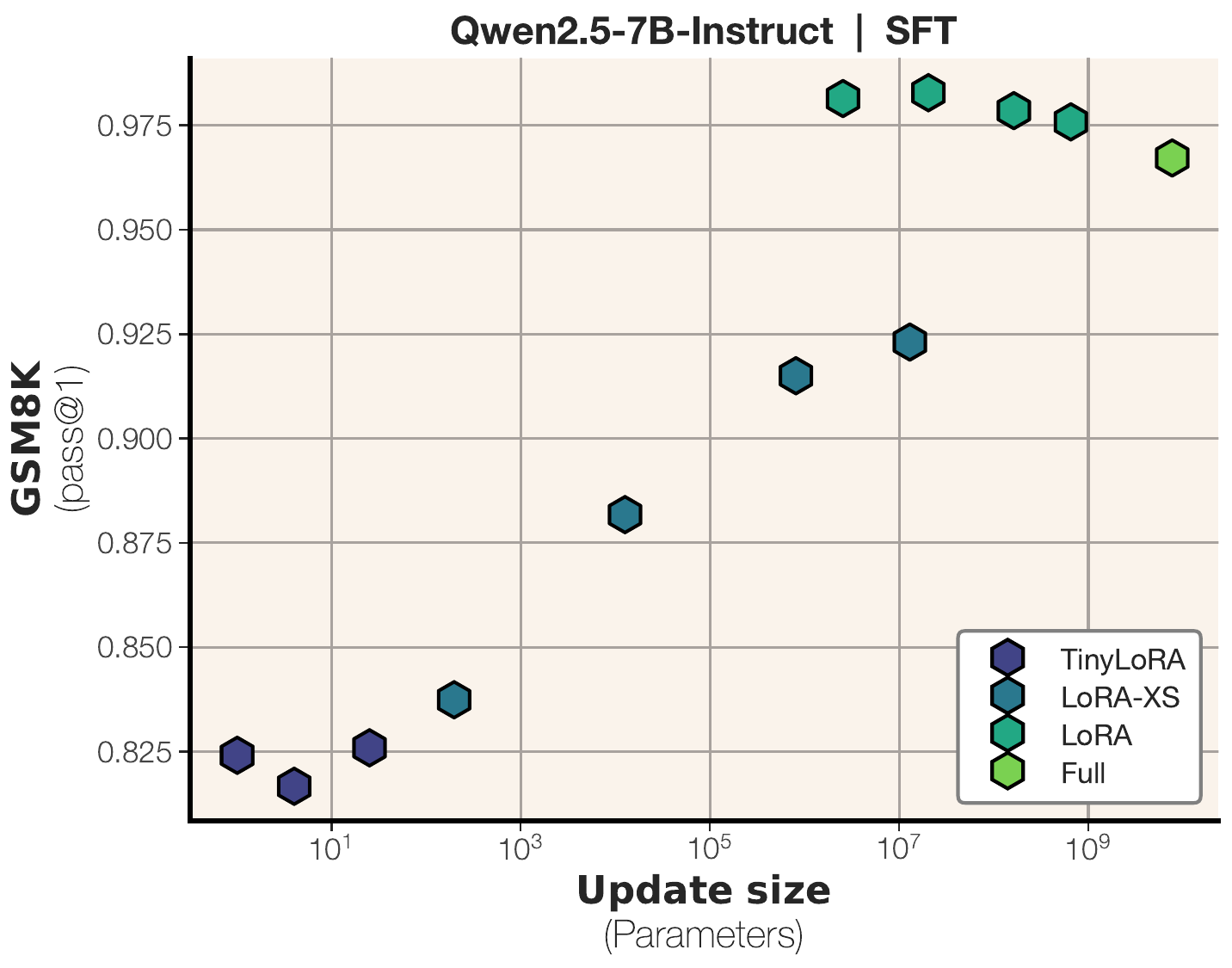}
        \caption{Using Qwen2.5-7B-Instruct as a base model, SFT works best with larger update sizes of at least $1M$ parameters.}
        \label{fig:main-gsm8k-sft}
    \end{minipage}
\end{figure}

\section{Introduction}

Modern language models are post-trained to "reason" in additional thinking tokens through reinforcement learning. Training for reasoning typically takes place through reinforcement learning with verifiable rewards (RLVR) and evaluated in reasoning-intensive domains such as math and coding \citep{openai2024openaio1card, shao2024deepseekmathpushinglimitsmathematical, deepscaler2025}.

Although many practitioners finetune the entire model for reasoning (typically in the billions of parameters) \citep{guha2025openthoughtsdatarecipesreasoning}, a popular class of parameter-efficient methods such as Low-Rank Adaptation (LoRA) \citep{hu2021loralowrankadaptationlarge} reduce this requirement from billions to just millions of "adapter" parameters, a reduction of several orders of magnitude. Even so, model updates remain large: for Llama3-8B, running LoRA at its smallest setting (rank 1) finetunes a minimum of 3M parameters \citep{schulman2025lora}. This parameter count looks surprisingly large compared to past work \citep{cuccu2019playingatarineurons}, which showed certain toy settings such as Atari games can be adequately solved using just six neurons.

There are many advantages to training with fewer parameters. Finetuning fewer parameters reduces per-GPU memory usage as well as the communication cost of distributed training. At inference time, more LoRAs can be stored in memory: a 10x reduction in the size of an adapter allows 10x more LoRAs to be served concurrently \citep{chen2023punicamultitenantloraserving}, enabling efficient personalization at scale. Past work \citep{biderman2024loralearnsforgets} also shows that low-rank model updates can mitigate forgetting by retaining important information present in the base model.

Given these benefits, how can we achieve even greater parameter efficiency? Prior work has focused primarily on supervised finetuning \citep{hu2021loralowrankadaptationlarge, li2021prefixtuningoptimizingcontinuousprompts, zaken2022bitfitsimpleparameterefficientfinetuning, biderman2024loralearnsforgets}. We observe that good performance in supervised learning requires absorbing many more bits of information into the model than in RL. Excitingly, the converse also holds: when finetuning with RL, we can achieve comparable performance with much smaller updates. This observation leads us to propose an RL-based alternative that enables finetuning with orders of magnitude fewer parameters than standard LoRA-based SFT.

We consider conventional LoRA \citep{hu2021loralowrankadaptationlarge} and the smaller LoRA-XS \citep{hu2021loralowrankadaptationlarge} which allow us to scale update sizes from $10^6$ down to $10^4$ parameters. On math tasks, RL is much more effective than SFT in the low-parameter regime (under 1 million parameters trained). On the GSM8K dataset of math word problems, Qwen-2.5-7B-Instruct improves from 76\% to 95\% while training just 10,000 parameters. Since models experience such little performance degradation at $10^4$ parameters, we develop a new method (TinyLoRA) to scale LoRA down arbitrarily: our smallest update size is just a single parameter. TinyLoRA is an extra low-rank variant of LoRA that scales smoothly down to just one trained parameter. When training Qwen2.5-7B-Instruct with TinyLoRA and GRPO on GSM8K, \textbf{we achieve 91\% accuracy while finetuning just 13 total parameters}, an update size of just 26 bytes.

We show that such low-capacity model updates only succeed in the case of RL, and are not information-dense enough to work well for SFT. On GSM8K, models trained with GRPO can obtain 90\% accuracy with fewer than 100 parameters, while models of this capacity trained with SFT barely outperform base models on the training set. These findings imply that RL makes fundamentally more information-dense updates than SFT and is more suitable for training in the tiny update size regime.

We also evaluate TinyLoRA on more sophisticated learning-to-reason benchmarks including MATH, AIME, and AMC. We again observe strong performance in the extremely-low parameter regime. As one datapoint, finetuning Qwen-2.5-7B-Instruct for just 196 parameters retains 87\% of its absolute performance improvement, averaged across six difficult math benchmarks.

\begin{figure}
    \centering
    \begin{minipage}{0.48\textwidth}
        \centering
        \includegraphics[width=\linewidth]{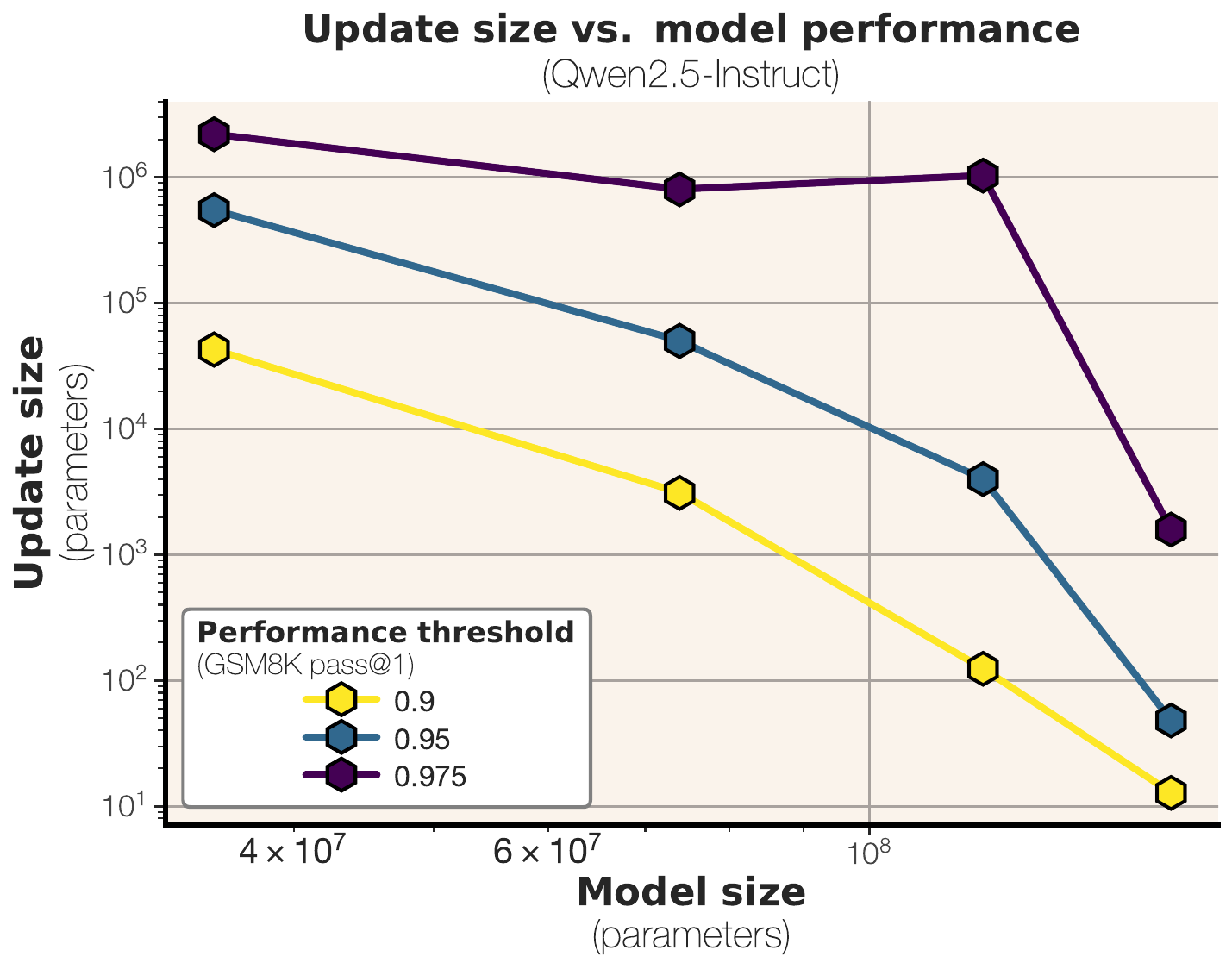}
        \caption{Minimal-sized parameter update to hit threshold of maximum performance vs. backbone model size. Larger models require smaller updates to reach e.g. 95\% of peak performance.}
        \label{fig:analysis-update-size}
    \end{minipage}
    \hfill
    \begin{minipage}{0.48\textwidth}
        \centering
        \includegraphics[width=\linewidth]{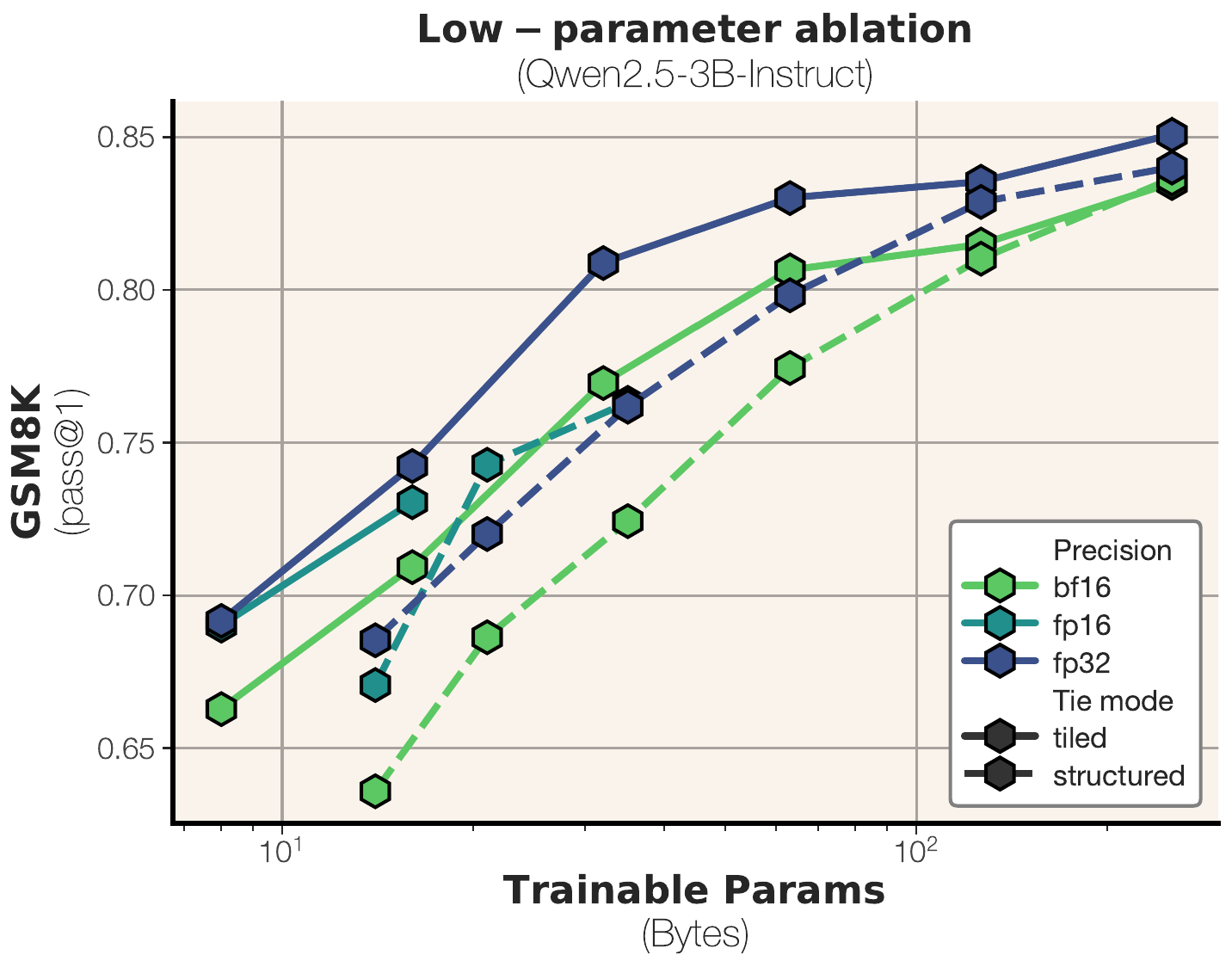}
        \caption{Performance ablation using Qwen2.5-3B-Instruct under extremely small update size (<1KB). Surprisingly, storing parameters in fp32 is most performant bit-for-bit.}
        \label{fig:analysis-tiling}
    \end{minipage}
\end{figure}

\section{Related Work}

\paragraph{Parameter-efficient finetuning.} Our progress scaling LoRA to low parameter counts is part of an ongoing effort to improve low-rank adaptation as evidenced through the slew of proposed related mesthods such as VeRA \citep{kopiczko2023vera}, VB-LoRA \citep{li2024vbloraextremeparameterefficient}, LoRA-XS \citep{bałazy2025loraxslowrankadaptationextremely}, UniLoRA \citep{li2025unilora}, AdaLoRA \citep{zhang2023adaloraadaptivebudgetallocation}, LoRA Drop \citep{zhou2024loradropefficientloraparameter}, NoRA \citep{lin2024noranestedlowrankadaptation}, WeightLoRA \citep{veprikov2025weightloranecessaryadapters}, and ShareLoRA \citep{song2025shareloraparameterefficientrobust}. We do not advocate for a specific initialization or even parameterization. Rather, we focus on simple methods for scaling to these techniques to low numbers of parameters (<10K), where the vast majority of other work on low-rank adaptation is done at the 10K–10M parameter scale. We show that even more efficient parameterizations are possible because we use larger models (billions of parameters) and train with RL as opposed to SFT.

\paragraph{Learning under data and parameter constraints.} Another line of work questions how much data is necessary for reasoning, finding that a thousand examples is sufficient for SFT \citep{ye2025limoreasoning} and, under certain settings, RL can learn with only one example \citep{wang2025reinforcementlearningreasoninglarge}. Our analysis of the pareto frontier between update size and performance relates to the general principle that overparameterized models lie on a manifold of a low \textit{intrinsic dimension} \citep{li2018measuringintrinsicdimensionobjective,aghajanyan2020intrinsicdimensionalityexplainseffectiveness}.

\paragraph{Comparing SFT and RL.} A line of recent work compares supervised finetuning with reasoning tokens (SFT) to reinforcement learning-based methods such as policy gradient.  
Our analysis of the relationship between data size and LoRA performance relates to finetuning scaling laws \citep{zhang2024scalingmeetsllmfinetuning} and studies of LoRA learning capacity \citep{biderman2024loralearnsforgets}. Some works argue that LoRA makes only superficial changes to models \citep{shuttleworth2025loravsfinetuningillusion}. Other work shows that the updates made by RL are small and relatively sparse \citep{mukherjee2025reinforcementlearningfinetunessmall}. Recently, a number of papers have shown that RL generalizes better out-of-domain than SFT \citep{chu2025sftmemorizesrlgeneralizes}, forgets less \citep{shenfeld2025rlsrazoronlinereinforcement}, and generally amplifies capabilities present in the base model rather than injecting new ones \citep{rajani2025scalpelvshammergrpo}. Concurrent work \citep{schulman2025lora} demonstrates LoRA can match full finetuning even at $r=1$ but does not try parameterizations smaller than $3M$.


\section{Update Capacity of SFT and RL}

\newcommand{\mcX}{\mathcal{X}}
\newcommand{\mcY}{\mathcal{Y}}
\newcommand{\mcR}{\mathcal{R}}

In this section we revisit two popular learning algorithms, supervised finetuning and reinforcement learning, and consider the amount of capacity it might take to train a model to good performance using either. Throughout, we consider a model $\pi_\theta$ that maps inputs $x \in \mcX$ (in this work, word problems) to outputs $y \in \mcY$ (solutions).

We study SFT with next-token prediction loss:
\[
\mathcal{L}_{\text{SFT}}(\theta) = -\mathbb{E}_{(x,y) \sim \mathcal{(X,Y)}} \left[ \sum_{t=1}^{|y|} \log \pi_\theta(y_t | x, y_{<t}) \right]
\]
and a class of reinforcement learning algorithms known as policy gradient:
\[
\nabla_\theta J(\theta) = \mathbb{E}_{x \sim \mathcal{X}; y\sim  \pi_\theta} \left[ \sum_{t=1}^{|y|} \nabla_\theta \log \pi_\theta(y_t | x, y_{<t}) \cdot R(y) \right]
\]

\paragraph{Information content of training data.}
Consider training a model $\theta$ on a sample $(x,y)$. The information content of $(x,y)$ for the model can be seen as the minimum description length of $(x,y)$ under the model \citep{rissanen1978mdl}. This quantity is independent of how $(x,y)$ is sampled.

For SFT, $(x,y)$ is sampled from a distribution $(\mcX, \mcY)$ for which the expected reward is high. For simplicity, assume the reward is binary and the underlying distribution is $(\mathcal{X,Y})$ where $\mathcal{Y} \equiv \pi_\theta(\mcX) \mid r(\mcX, \pi_\theta(\mathcal{X}))=1$.

For policy gradient methods, the observed data is a batch of continuations sampled from $(\mcX, (\mcY, \mcR)^k)$ where $\mcY\equiv \pi_\theta(\mcX)$ and $\mcR \equiv r(\mcX, \mcY)$. These samples are regenerated at each epoch using the updated model.

\paragraph{RL presents more data but less information.}
In one sense, the information presented during RL appears greater than SFT: at each epoch and for each prompt $x$, we observe $k$ fresh continuations with new entropy, while SFT uses only one continuation repeated across epochs. However, most of this information is noise. Without the reward annotation, the continuations $\mcY$ contain no useful signal for improving task performance. The relevant information lies entirely in the reward, which is only $k \cdot H(\mcR)$ bits per prompt---bounded by $k$ when the reward is binary.

\paragraph{Signal separation.}
Despite this sparsity of useful information in RL, the signal is cleanly separated from noise. Reward-relevant features correlate with $r$, reward-irrelevant features do not. Resampling amplifies this separation—the correlated signal accumulates while uncorrelated variation cancels.

In SFT, however, the training signal is a demonstration $y$ with no reward annotation. The model cannot distinguish which features of $y$ are task-relevant. Without a mechanism to separate signal from noise, SFT must treat all tokens as equally informative, storing both the useful structure and the irrelevant details.

\paragraph{Hypothesis.}

We hypothesize that SFT is less capable in low-parameter regimes because minimizing its objective requires a model to absorbs many bits of information, only a fraction of which are relevant to task performance. RL receives a sparser, cleaner signal, allowing it to learn effectively with less capacity.

\section{Parameter-Efficient Finetuning with TinyLoRA}

The low capacity requirements of reinforcement learning lead us to question just how small an adapter might be while still providing adequate performance. Recent work \citep{schulman2025lora, yin2025evaluatingparameterefficientmethods} has observed that LoRA can be effective at update sizes as small as $10^6$ parameters. In this section we build upon prior work LoRA and LoRA-XS to propose TinyLoRA, a method for parameter-efficient updates that scale down to a single trained parameter.

\subsubsection{LoRA and LoRA-XS}
Low-rank adaptation, or LoRA \citep{hu2021loralowrankadaptationlarge}, adapts a frozen linear layer $W \in \mathbb{R}^{d \times k}$ with a low-rank update:
\[
W' = W + AB
\]
where $A \in \mathbb{R}^{d \times r}$ and $B \in \mathbb{R}^{r \times k}$ are trainable, and $W$ remains frozen. The number of trainable parameters scales as $\mathcal{O}(dr)$ per module. Applying LoRA to $m$ modules across $n$ layers yields $\mathcal{O}(nmdr)$ total parameters—typically millions for billion-parameter models \citep{biderman2024loralearnsforgets, schulman2025lora}.
LoRA-XS \citep{bałazy2025loraxslowrankadaptationextremely} reduces the per-module parameter count from $\mathcal{O}(dr)$ to $\mathcal{O}(r^2)$:
\[
W' = W + U \Sigma R V^\top
\]
where $U \in \mathbb{R}^{d \times r}$, $\Sigma \in \mathbb{R}^{r \times r}$, and $V \in \mathbb{R}^{k \times r}$ are from the truncated SVD of $W$, and only $R \in \mathbb{R}^{r \times r}$ is trainable. This can be viewed as learning to recombine the dominant singular directions of $W$, and outperforms randomly-initialized LoRA in practice.

\subsubsection{Reducing the number of trainable parameters with TinyLoRA}

Since the size of a LoRA is inherently tied to the model width, we propose a new technique for scaling down the number of trained parameters. Our technique allows us to train models with as few as $1$ changed parameter.

\paragraph{Reducing the size of $R$.} Even with $r=1$, LoRA-XS requires at least one parameter per adapted module. We reduce this further by replacing the $r \times r$ matrix $R$ with a low-dimensional trainable vector $\mathbf{v} \in \mathbb{R}^{u}$ projected through a fixed random tensor $P \in \mathbb{R}^{u \times r \times r}$. The update rule for TinyLoRA is:
\[
W' = W + U \Sigma \left(\sum_{i=1}^{u} v_i P_i\right) V^\top
\]
where $U, \Sigma, V$ are from the truncated SVD of $W$, and $P_i \in \mathbb{R}^{r \times r}$ are fixed random matrices. Each module trains only $u$ parameters. With weight tying across $m$ modules in $n$ layers, the total trainable parameters scale as $\mathcal{O}(nmu / n_{\text{tie}})$, reducing to a single parameter when all modules share weights.

\paragraph{Parameter sharing.}
\begin{table}[t!]
\centering
\begin{tabular}{@{}lcc@{}}
\toprule
\textbf{Method} & \textbf{Trainable Parameters} & \textbf{Minimum} \\
\midrule
FT & $\mathcal{O}(nmd^2)$ & $nmd^2$ \\
LoRA & $\mathcal{O}(nmdr)$ & $2nmd$ \\
LoRA-XS & $\mathcal{O}(nmr^2)$ & $nm$ \\
VeRA & $\mathcal{O}(nm(d + r))$ & $2nmd$ \\
TinyLoRA & $\mathcal{O}(nmu)$ & $1$ \\
\bottomrule
\end{tabular}
\caption{Parameter usage comparison per-layer with $m$ adapted modules per layer, model width $d$, rank $r$, and TinyLoRA projection dimension $u$. With weight tying across modules, TinyLoRA can reduce to a single trainable parameter.}
\label{tab:parameter_usage}
\end{table}
Prior work \citep{dettmers2023qloraefficientfinetuningquantized,biderman2024loralearnsforgets} showed that LoRA performs best when applied to both MLP and attention modules. In a typical transformer architecture such as LLaMA-3 \citep{llama3model}, LoRA is applied seven times per block: to the query, key, value, and output projections in self-attention, and to the up, down, and gate projections in the MLP. 

For a model like LLaMA-3 70B with 80 layers, even the minimal case of $u=1$ (or $r=1$ for LoRA-XS) requires $80 \times 7 = 560$ trainable parameters. We reduce parameter count further by sharing the trainable vector $\mathbf{v}$ across modules. We define the weight tying factor $n_{\text{tie}}$ as the number of modules sharing a single $\mathbf{v}$, yielding $nm u / n_{\text{tie}}$ total trainable parameters for $n$ layers and $m$ modules per layer. With full weight tying ($n_{\text{tie}} = nm$), all modules share a single $\mathbf{v}$, reducing the total to just $u$ parameters—as few as one. Differences between TinyLoRA and related methods are shown in \Cref{tab:parameter_usage}.

\section{Experiments}

\subsection{Design}

The goal of our experiments is to improve math reasoning in language models while minimizing the number of trained parameters. We evaluate on GSM8K \citep{cobbe2021trainingverifierssolvemath}, a dataset of 7,500 math word problems, and on the harder MATH training set \citep{hendrycks2021measuringmathematicalproblemsolving}.

We consider model training via either \textit{supervised finetuning} (SFT), i.e.\ next-token prediction, or \textit{reinforcement learning}, in particular Group Relative Policy Optimization (GRPO) \citep{shao2024deepseekmathpushinglimitsmathematical}. We conduct experiments on instruction-tuned language models from the Llama-3 \citep{llama3model} and Qwen-2.5 \citep{qwen2025qwen25technicalreport} families.

We run four baselines: full finetuning, LoRA, LoRA-XS, and TinyLoRA. We test ranks $\{1, 8, 64, 256\}$ for all LoRA variants; for TinyLoRA, we consider sharing $\{1, 8, 64, 256\}$ layers. All our RL experiments use exact-match reward.

For GSM8K, we do not use any KL penalty, and we train on the full dataset for three epochs with $4$ samples per problem and a batch size of $64$. We sample with a maximum generation length of $4096$.

For MATH training, we follow the settings of SimpleRL \citep{zeng2025simplerlzooinvestigatingtamingzero}, which includes a larger training dataset containing subsets such as GSM8K and MATH. The data is partitioned by difficulty level; we only train on the hardest difficulty, which consists of 8,523 problems. We generate data with a maximum prompt length of $1024$ and response length of $3072$ tokens, and use the `boxed' chat template. We keep all other hyperparameter settings of SimpleRL: KL coefficient $0.001$, temperature $1.0$, batch size $256$, and $8$ generations per response.

Since changes in update size are known to alter effective learning rate \citep{biderman2024loralearnsforgets,schulman2025lora}, we sweep over learning rates $\{10^{-7}, 5 \times 10^{-7}, 10^{-6}, 5 \times 10^{-6}, 10^{-5}, 10^{-4}, 2 \times 10^{-4}\}$ and take the top-performing learning rate at each update size, averaged over three random seeds.

\paragraph{Evaluation.} For GSM8K-only training, we evaluate on the GSM8K validation set. For MATH experiments, we additionally evaluate on the standard datasets proposed by SimpleRL \citep{zeng2025simplerlzooinvestigatingtamingzero}: MATH500 \citep{hendrycks2021measuringmathematicalproblemsolving}, Minerva \citep{lewkowycz2022solvingquantitativereasoningproblems}, GAOKAO \citep{zhang2024gaokao}, OlympiadBench \citep{he2024olympiadbench}, CollegeMath \citep{feynman2024mathscollege}, AIME 24 \citep{li2024numinamath}, and AMC23 \citep{he2024olympiadbench}. All datasets represent math word problems of varying degrees of difficulty and come from the canonical Qwen-Math evaluation \citep{qwen2025qwen25technicalreport}.

\paragraph{Implementing TinyLoRA in vLLM.} We run all RL experiments within the open-source VERL framework \citep{sheng2024hybridflow}, using vLLM \citep{kwon2023efficientmemorymanagementlarge} for inference. However, vLLM requires custom kernels for LoRA and only supports a minimum LoRA rank of 4 (and no LoRA variants). We circumvent this issue by modifying training to use the \textit{merged} model weights for inference, and the true LoRA model only for the final forward pass. This creates a natural numerical mismatch between training and inference, which we mitigate via truncated importance sampling \citep{ionides2008TruncatedIS, yao2025offpolicy}. This approach allows us to test novel parameter-efficient methods without implementing each one in separate low-level code.

\section{Results}

\subsection{Initial low-parameter results: TinyLoRA, LoRA-XS, and LoRA}

We first train and evaluate Qwen2.5-7B-Instruct on GSM8K with GRPO, with results shown in \Cref{fig:main-gsm8k-grpo}. We observe a relatively smooth interpolation from TinyLoRA ($1$–$10^3$ parameters trained) to LoRA-XS ($10^3$–$10^6$ parameters trained) to LoRA ($>10^6$ parameters trained.) We are able to recover $95\%$ of the net performance improvement on GSM8K with only $120$ parameters trained. \textbf{With TinyLoRA, we see a 4\% performance increase when training just a single parameter.}

\subsection{SFT vs. RL}

We display full sweep results for RL (\Cref{fig:main-gsm8k-grpo}) vs. SFT (\Cref{fig:main-gsm8k-sft}). We find RL is significantly more efficient at low parameter counts:  for a baseline of 76\%, models trained with RL score 91\% (15\% absolute improvement) at 13 parameters and 95\% at 120 parameters.  Models trained with SFT do much worse: 83\% at 13 parameters and 84\% at 120.

We also observe a less-smooth transition from LoRA-XS at high rank to LoRA at low rank. We note that SFT training directly on answers is a version of \textit{off-policy} learning, and dynamics may change when training directly on model outputs \citep{mukherjee2025reinforcementlearningfinetunessmall, chen2025retainingdoingroleonpolicy}.

\subsection{Training on MATH}

\definecolor{metabg}{HTML}{F1F4F7}

\begin{table}[t]
\centering
\setlength{\tabcolsep}{6pt}
\renewcommand{\arraystretch}{1.15}
\begin{tabular}{llrrrrrrr}
\toprule
\# & GSM8K & MATH 500 & Minerva Math & Olympiad Bench & AIME24 & AMC23 & Avg. \\
\midrule
\multicolumn{8}{>{\columncolor{metabg}}c}{\textbf{Qwen2.5-3B-Instruct}} \\
(0) & 76.0 & 55.0 & 18.5 & 21.3 & 2.1 & 23.4 & 32.7 \\
16 & 80.9 & 64.0 & 19.9 & 23.0 & 3.0 & 31.5 & 37.1 \\
63 & 85.3 & 64.1 & 20.1 & 26.6 & 7.3 & 36.0 & 39.9 \\
 252 & 85.4 & 66.4 & 28.3 & 29.3 & 13.3 & 47.5 & 45.0 \\
 504 & 86.1 & 66.6 & 28.7 & 30.8 & 16.7 & 47.5 & 46.1 \\
 8,064 & 87.2 & 67.8 & 28.3 & 30.7 & 10.0 & 47.5 & 45.2 \\
 129,024 & 86.7 & 67.8 & 29.4 & 32.3 & 10.0 & 55.0 & 46.9 \\
 (3,085,846,528) & 87.0 & 69.0 & 31.7 & 33.1 & 15.0 & 52.2 & 48.0 \\
\midrule
\multicolumn{8}{>{\columncolor{metabg}}c}{\textbf{Qwen2.5-7B-Instruct}} \\
(0) & 88.2 & 64.6 & 25.7 & 30.1 & 3.3 & 30.0 & 40.3 \\
13 & 91.8 & 74.6 & 27.1 & 36.3 & 16.0 & 54.5 & 50.1 \\
49 & 91.5 & 74.2 & 26.6 & 37.2 & 12.6 & 55.5 & 49.6 \\
196 & 92.2 & 76.6 & 37.1 & 38.8 & 16.7 & 57.5 & 53.2 \\
392 & 92.2 & 77.0 & 35.7 & 40.1 & 16.7 & 65.0 & 54.4 \\
6,272 & 91.9 & 78.0 & 37.5 & 41.0 & 16.7 & 57.5 & 53.8 \\
100,352 & 92.8 & 78.0 & 37.1 & 43.3 & 16.7 & 60.0 & 54.6 \\
 (7,615,487,488) & 91.7 & 78.2 & 38.6 & 40.4 & 20.0 & 62.5 & 55.2 \\
\midrule
\multicolumn{8}{>{\columncolor{metabg}}c}{\textbf{Qwen2.5-7B-Math}} \\
(0) & 65.5 & 63.6 & 12.5 & 25.8 & 13.3 & 42.5 & 37.2 \\
196 & 72.5 & 62.0 & 26.5 & 31.4 & 20.0 & 50.0 & 43.7 \\
392 & 86.0 & 74.8 & 31.6 & 37.9 & 26.7 & 60.0 & 52.8 \\
6,272 & 87.0 & 77.4 & 28.7 & 40.0 & 16.7 & 67.5 & 52.9 \\
100,352 & 87.0 & 78.6 & 32.7 & 39.0 & 30.0 & 62.5 & 55.0 \\
 (7,615,487,488) & 90.2 & 80.2 & 37.5 & 39.0 & 40.0 & 70.0 & 59.5 \\
\bottomrule
\end{tabular}

\caption{Performance on math reasoning using Qwen2.5 models. (0) indicates the base model without training; (*) indicates full finetuning (no LoRA). All experiments use the GRPO algorithm and train for three epochs. Full and baseline results are reported from \citep{zeng2025simplerlzooinvestigatingtamingzero}.}
\label{tab:math}
\end{table}

We investigate the performance of low-parameter adaptation methods under the SimpleRL training framework \citep{zeng2025simplerlzooinvestigatingtamingzero}. We train models across update sizes for the 7B Qwen model and evaluate on math reasoning tasks; results are shown in \Cref{tab:math}. We consider updates of \{100, 10,000, 1M\} parameters as compared to full finetuning.

\begin{figure}
    \centering
    \includegraphics[width=0.95\linewidth]{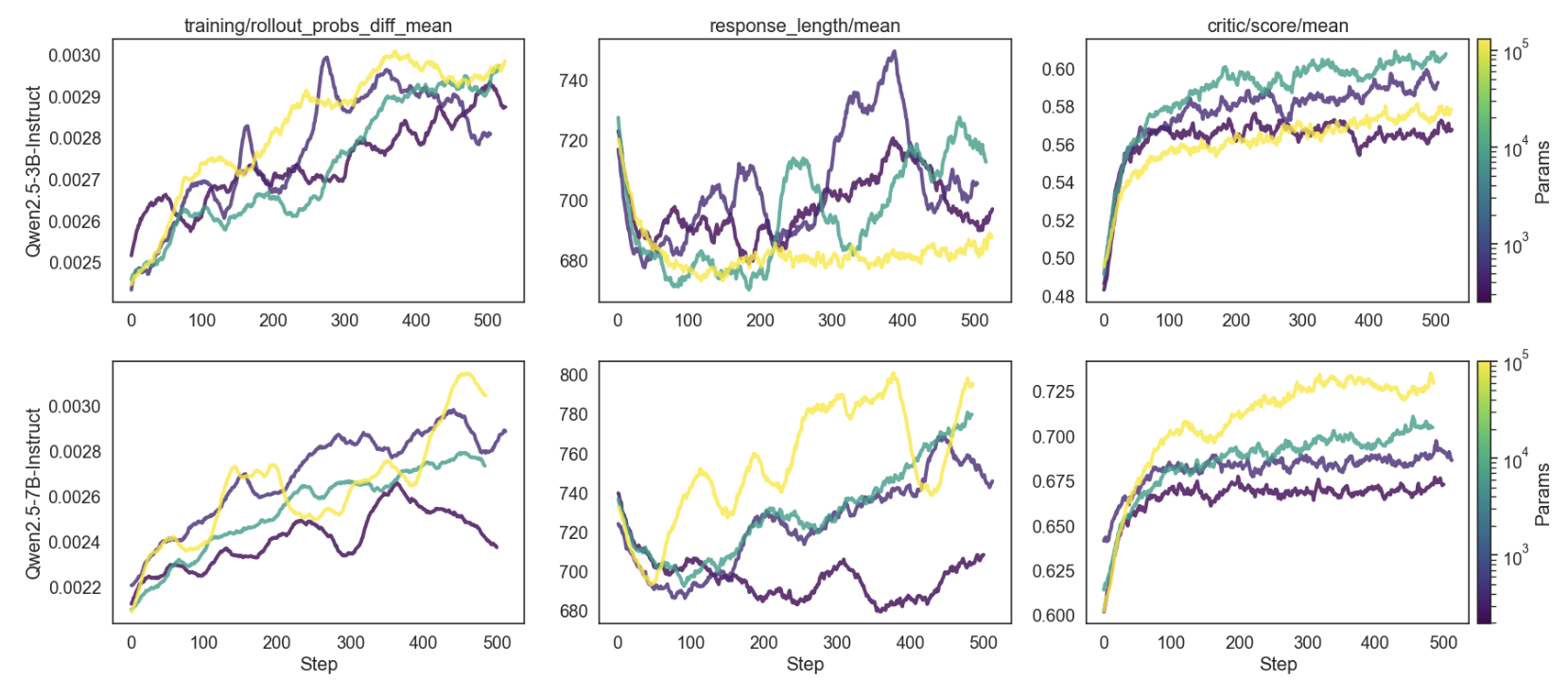}
    \caption{TinyLoRA performance during training on MATH.}
    \label{fig:math-training}
\end{figure}

We also visualize metrics across training steps (\Cref{fig:math-training}). We visualize the mean reward and mean response length. Larger parameter counts tend to achieve higher rewards and generate longer responses, but all parameter counts are getting some rewards – even update sizes as small as 16 parameters. Visualizing the KL divergence between the log probabilities during training and inference shows negligible KL between the train and inference models, showing our technique of merging the LoRA weights at each training step is working properly.

\subsection{Scaling across model backbones}
We then scale TinyLoRA from the XS $r=1$ case ($np$ parameters) down to a single trained parameter, for Qwen3-8B on GSM8K. Results are shown in \Cref{fig:main-gsm8k-grpo}. For the task of GSM8K, Qwen is more parameter-efficient than LLAMA at every scale. In an absolute sense, Qwen is surprisingly efficient, achieving 94.7\% accuracy with only 13 parameters trained. Qwen even achieves decent accuracy at 1 parameter, scoring near 82\%. (around a 5\% improvement over the baseline). This finding may corroborate recent observations that Qwen in particular may have exposure to similar examples during its pretraining \citep{wu2025reasoningmemorizationunreliableresults}.

Even in the case of LoRA-XS, which enables adapting layers with a single parameter, we are upper-bounded by the total number of linear layers in the model. Large models can contain hundreds of linear layers, which would allow for hundreds of bytes of storage at $r=1$. Empirically, we observe that Qwen can still learn nearly as well with only a few hundred parameters.

Although LLAMA has a similar upper-bound (from full finetuning), in this low-parameter setting LLAMA is less responsive than Qwen and reaches 85\% with an update size of 1KB (500 parameters trained in bf16). Unlike Qwen, when we train fewer than five parameters, LLAMA barely improves performance above baseline.

\paragraph{Ranks 1 to 128.} We visualize performance from $r=1$ to $r=128$ for Qwen and LLAMA in \Cref{fig:main-gsm8k-grpo}. This covers update sizes from 1KB (r=1) to 8MB (r=128). We see consistent monotonic increase in performance, suggesting that although the performance boost is decreasing per-parameter, finetuning additional parameters does help up to an update size of around 2MB.

\begin{figure}
    \centering
    \includegraphics[width=0.95\linewidth]{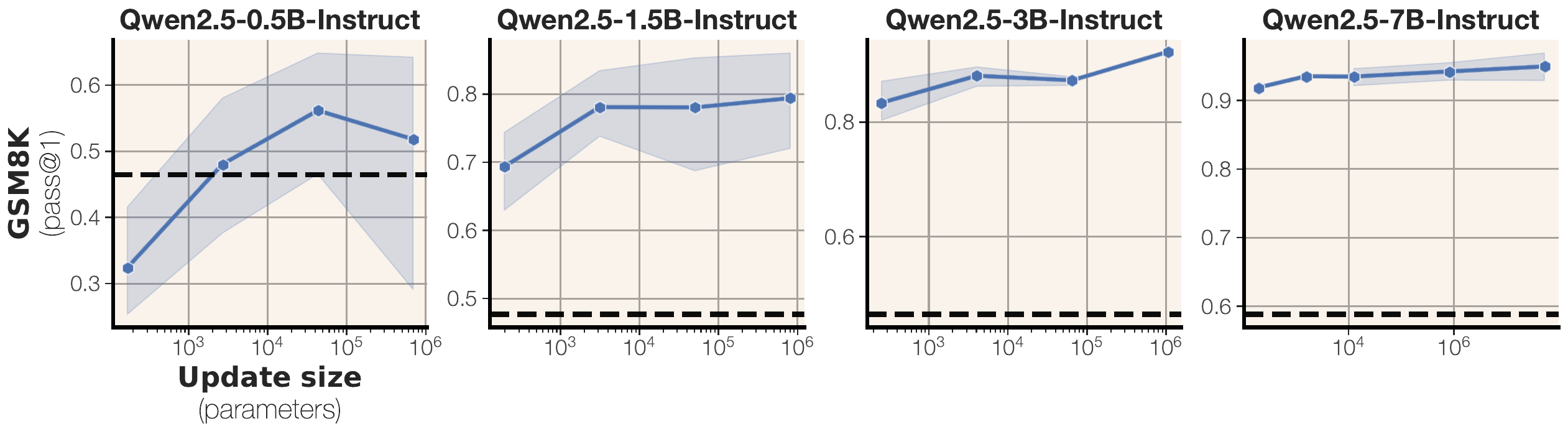}
    \caption{TinyLoRA performance (on GSM8K GRPO) across backbone size, with no-training baselines as dashed lines. Small updates only improve larger models.}
    \label{fig:lora-loraxs-scaling}
\end{figure}

\paragraph{From LoRA-XS to LoRA, across model backbones.} We display results with LoRA and with LoRA-XS in \Cref{fig:lora-loraxs-scaling} across backbone sizes. Our results are mixed: LoRA-XS clearly outperforms LoRA on the smallest model, corroborating findings of \citep{bałazy2025loraxslowrankadaptationextremely}. But as model size grows, this trend becomes less pronounced, and degrades to performance scaling with the number of trained parameters. (We note that the number of LoRA modules increases significantly in larger models, and this may make all the difference here.)

In general, low-parameter adaptation works better with larger models – in such cases, smaller adapters appear closer to the performance ceiling. We plot the update size vs. relative performance increase across models from the Qwen2.5-Instruct family in \Cref{fig:analysis-update-size}. The trend is extremely clear: as model size grows, the model is programmable to within $95\%$ of its full finetuning performance with fewer \textit{absolute} parameters. These trends indicate that extremely large (trillion-scale) models may be easily trainable for many tasks w ith just a handful of parameters.

\subsection{Scaling in the bit-constrained regime}
We question how to scale TinyLoRA when the parameter count is small (tens or hundreds of parameters) and the constraint is on the total update size in \textit{bytes}. This scenario is common in distributed training, where communicating model parameter updates can be a major bottleneck. We consider two parameter-sharing strategies: \textit{structured}, where nearby modules of the same type share parameters, and \textit{tiled}, where nearby modules of similar depth share parameters, agnostic of type. 

Results across sharing strategies and precisions are shown in \Cref{fig:analysis-tiling}. Surprisingly, tiling outperforms structured sharing: we see no benefit from sharing parameters between e.g. query projection modules. With all-layer sharing and float16 precision, Qwen still trains to a level of 70\% on GSM8K, an absolute improvement of over 10\%. We also notice that fp32 outperforms bf16 and float16 even when accounting for its twice-as-large update size in bytes.

\section{Ablations}
We train a number of smaller models to analyze the effect of TinyLoRA hyperparameters. Using GRPO, we finetune Qwen-2.5-3B-Instruct and LLaMA-3.2-3B-Instruct, sweeping over the frozen SVD rank $r$, the trainable projection dimension $u$, and the weight tying factor $n_{\text{tie}}$ (the number of modules sharing a single $\mathbf{v}$).

\begin{figure}[h]
    \centering
    \includegraphics[width=.96\textwidth]{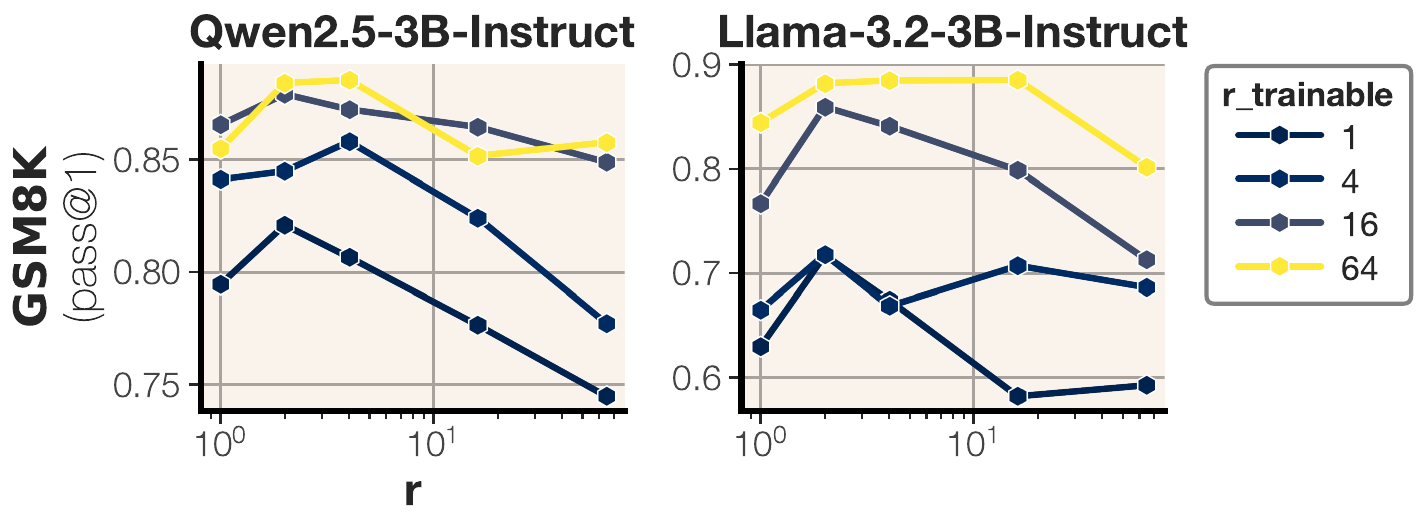}
    \caption{TinyLoRA performance across frozen rank $r$ and trainable rank $r_{trainable}$. }
    \label{fig6:ablation-r-trainable}
\end{figure}

\paragraph{Selection of frozen rank $r$.} We first analyze performance as a function of the frozen rank $r$ (\Cref{fig6:ablation-r-trainable}). While one might expect that preserving more singular directions would improve expressivity, we find diminishing returns: increasing $r$ from 1 to 2 yields modest gains, but larger values degrade performance. We hypothesize that higher ranks introduce more degrees of freedom in the frozen $U$, $\Sigma$, $V$ components, making optimization of the small trainable vector $\mathbf{v}$ more difficult. Based on these results, we use $r=2$ for all main experiments.

\begin{figure}[h]
    \centering
    \includegraphics[width=.96\textwidth]{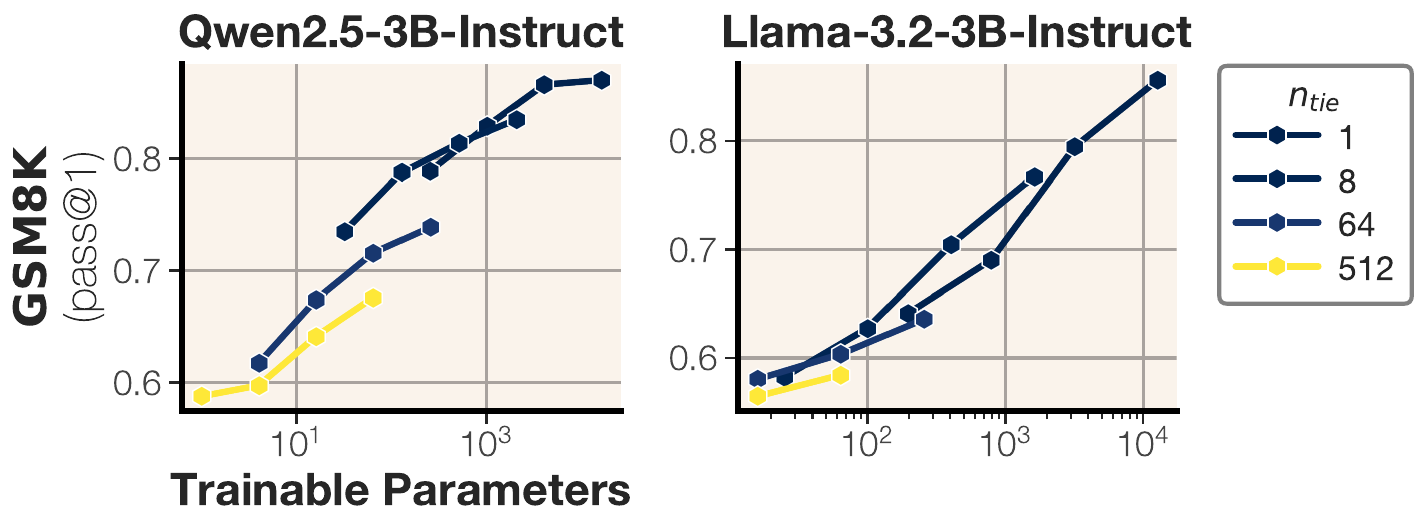}
    \caption{Performance ablation with Qwen2.5-3B-Instruct across numbers of tied layers and trainable ranks.}
    \label{fig6:ablation-n-tie}
\end{figure}

\paragraph{Trading off $u$ and $n_{\text{tie}}$.} For a fixed parameter budget, TinyLoRA requires balancing the per-module projection dimension $u$ against the weight tying factor $n_{\text{tie}}$. Increasing $u$ adds expressivity per module but increases parameter count; increasing $n_{\text{tie}}$ reduces parameters but forces modules to share the same $\mathbf{v}$. We plot this tradeoff in \Cref{fig6:ablation-n-tie}. Performance generally improves with larger $u$ and smaller $n_{\text{tie}}$—that is, more expressive per-module updates with less sharing. This suggests a clear guideline for practitioners: exhaust the $u$ budget (down to $u=1$) before increasing $n_{\text{tie}}$.

\section{Discussion}

Our work raises interesting questions about how model update size during finetuning scales with model size and by learning objective.

\textbf{Scaling trends.} Our findings indicate that LoRA becomes \textit{more effective} at smaller parameter counts as model size scales. Given a fixed dataset, larger models can be controlled with fewer parameters. This trend indicates that extremely parameter-efficient methods such as TinyLoRA will only grow in popularity as models continue to scale.

\textbf{How is this possible?} One might ask how it's possible to learn to solve a difficult task such as GSM8K in as few as 13 parameters. One theory is that the \textit{knowledge} required to solve the task is already stored in the parameters of the model, and only the \textit{style} has to change for task success. In particular, learning to generate longer outputs may be possible in few parameters, and has been shown to greatly improve performance on math and reasoning tasks \citep{shao2024deepseekmathpushinglimitsmathematical}.

\textbf{Peculiarities of Qwen and LLaMa.} We also observe that the Qwen-2.5 models are generally much more performant at small update sizes than LLaMA-3. In general, Qwen models of the same family require around 10x fewer parameters updated to reach equivalent performance to LLaMa. This may relate to differences in details in the architecture or pretraining processes of the models.

\textbf{Limitations of math datasets.} We note that our findings are limited to math datasets. Many results seem to indicate that, using reinforcement learning, very small updates (on the order of hundreds of bytes) are enough to facilitate learning to reason with reasonable performance. But these findings are limited to the scope of math-style reasoning, and may or may not generalize to other fields such as science or creative writing. We focus on math as it is the most popular application of reasoning models as of 2025, but look forward to seeing our findings extended and replicated in other domains.


\section{Conclusion}

We propose TinyLoRA and show that effective models can be tuned with far fewer parameters than previously thought to be necessary. Updates from TinyLoRA often come close to the performance of full finetuning can be expressed in very small file sizes – often less than 1KB in total. Our results show that some models can learn to do certain tasks while absorbing very small numbers of bytes, which calls into question what is actually learned during RL with verification. Finally, we show that this is only possible in the case of RL on large models, and does not work as well when not doing RL or not using large models.

\section{Acknowledgements}

Thanks to Jessy Lin, Dan Biderman, Sasha Rush, John Schulman, and Kamalika Chaudhuri for helpful feedback on this work.

\bibliographystyle{plainnat}
\bibliography{main}

\appendix
\section{Appendix}

\subsection{Additional ablations}

\begin{figure}[h]
    \centering
    \includegraphics[width=.96\textwidth]{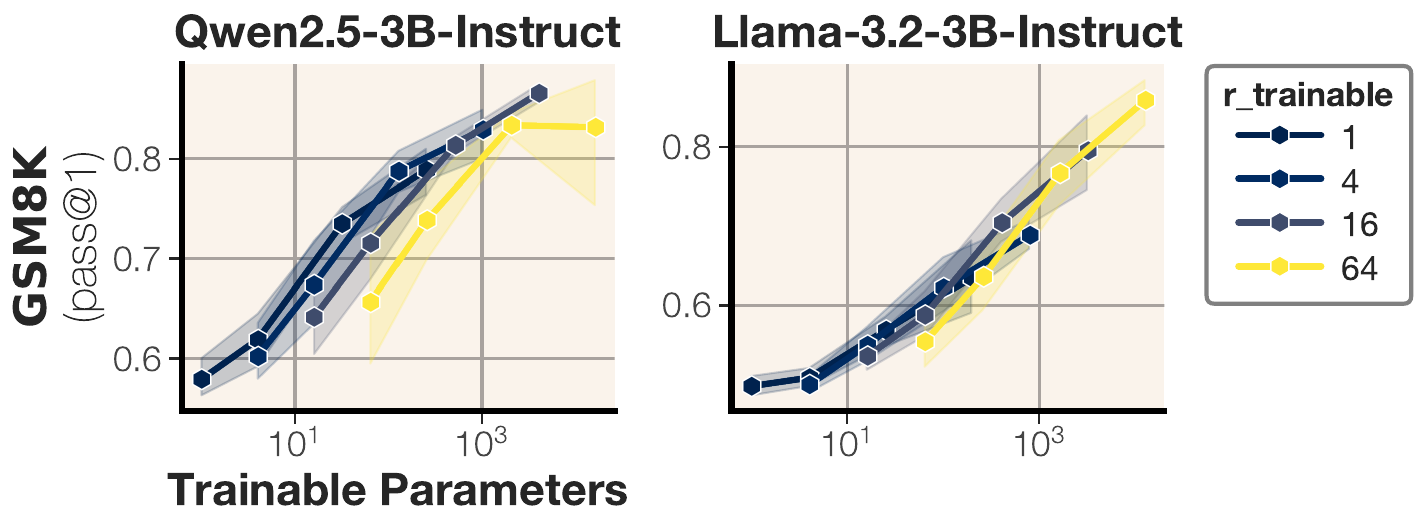}
    \caption{Performance ablation for Qwen2.5-3B-Instruct with different numbers of tied layers and ranks $r$.}
\end{figure}

We ran a number of ablations to isolate the right way to `spend' trainable parameters: is it better to have more, shared parameters or fewer-but-unique parameters? We sweep across values of $r_\text{trainable}$ for 3B versions of Qwen-2.5 and LLAMA-3.2 on GSM8K. Surprisingly, we find higher performance at matched parameter count for a \textit{lower} value of $r_\text{trainable}$, indicating parameters are better spent on fewer, unique parameters. We also observe again that Qwen far outperforms Llama at a constant update size: when training just 100 parameters, Llama achieves ~60\% accuracy, while Qwen averages 74\% across 3 random seeds.

\end{document}